\documentclass[10pt,twocolumn,letterpaper]{article}

\usepackage{iccv}
\usepackage{times}
\usepackage{epsfig}
\usepackage{graphicx}
\usepackage{amsmath}
\usepackage{amssymb}
\usepackage{mathtools}
\usepackage{bbm}
\usepackage{makecell, verbatim, sidecap}

\usepackage[breaklinks=true,bookmarks=false]{hyperref}

\iccvfinalcopy %

\usepackage{color}

\def\block1{{$\sf block1$}}
\def\block2{{$\sf block2$}}
\def\block3{{$\sf block3$}}
\def\block4{{$\sf block4$}}

\def\R{{\mathbb R}}

\DeclarePairedDelimiter{\floor}{\lfloor}{\rfloor}

\def\centerness{ {  \rm centerness  } }

\usepackage[margin=4pt,font=small,labelfont=bf,labelsep=endash,tableposition=top]{caption}

\renewcommand{\texttt}[1]{ $ {{\tt #1} } $}

\usepackage{xspace}

\renewcommand{\vec}[1]{\ensuremath{\pmb{#1}}}
\newcommand{\mat}[1]{\ensuremath{\mathbf{#1}}}
\newcommand{\set}[1]{\ensuremath{\mathscr{#1}}}

\makeatletter
\@tfor\next:=abcdefghijklmnopqrstuvwxyxz\do
 {\begingroup\edef\x{\endgroup
    \noexpand\@namedef{v\next}{\noexpand\vec{\next}}%
  }\x}
\@tfor\next:=ABCDEFGHIJKLMNOPQRSTUVWXYZ\do
 {\begingroup\edef\x{\endgroup
    \noexpand\@namedef{m\next}{\noexpand\mat{\next}}%
  }\x}
\@tfor\next:=ABCDEFGHIJKLMNOPQRSTUVWXYZ\do
 {\begingroup\edef\x{\endgroup
    \noexpand\@namedef{s\next}{\noexpand\set{\next}}%
  }\x}
\makeatother

\def\Ours{{FCNDet}\xspace}
\def\Ours{{FCOS}\xspace}
\def\eg{{\it e.g.}\xspace}

\begin{document}

\title{\Ours: Fully Convolutional One-Stage Object Detection}

\author{
Zhi Tian ~ ~ ~ ~  ~
Chunhua Shen\thanks{Corresponding author, email: \texttt{chunhua.shen@adelaide.edu.au}
                    }
        ~ ~ ~ ~ ~ 
Hao Chen ~ ~ ~ ~ ~ 
Tong He
\\
The University of Adelaide, Australia 
}

\maketitle
\thispagestyle{empty}

\begin{abstract}
We propose a fully convolutional one-stage object detector (FCOS) to solve object detection in a per-pixel prediction fashion, analogue to semantic segmentation. Almost all state-of-the-art object detectors such as RetinaNet, SSD,  YOLOv3, and Faster R-CNN rely on pre-defined anchor boxes. In contrast, our proposed detector FCOS is anchor box free, as well as proposal free. By eliminating the pre-defined set of anchor boxes, FCOS completely avoids the complicated computation related to anchor boxes such as calculating overlapping during training. More importantly, we also avoid all hyper-parameters related to anchor boxes, which are often very sensitive to the final detection performance. With the only post-processing non-maximum suppression (NMS), FCOS with ResNeXt-64x4d-101 achieves $44.7\%$ in AP with single-model and single-scale testing, surpassing previous one-stage detectors with the advantage of being much simpler. For the first time, we demonstrate a much simpler and flexible detection framework achieving improved detection accuracy. We hope that the proposed FCOS framework can serve as a simple and strong alternative for many other instance-level tasks.
Code is available at:

\href{https://tinyurl.com/FCOSv1}{{ $\tt tinyurl.com/FCOSv1$}}

\end{abstract}

\section{Introduction}
\begin{figure}[t!]
  \centering
  \includegraphics[width=\linewidth]{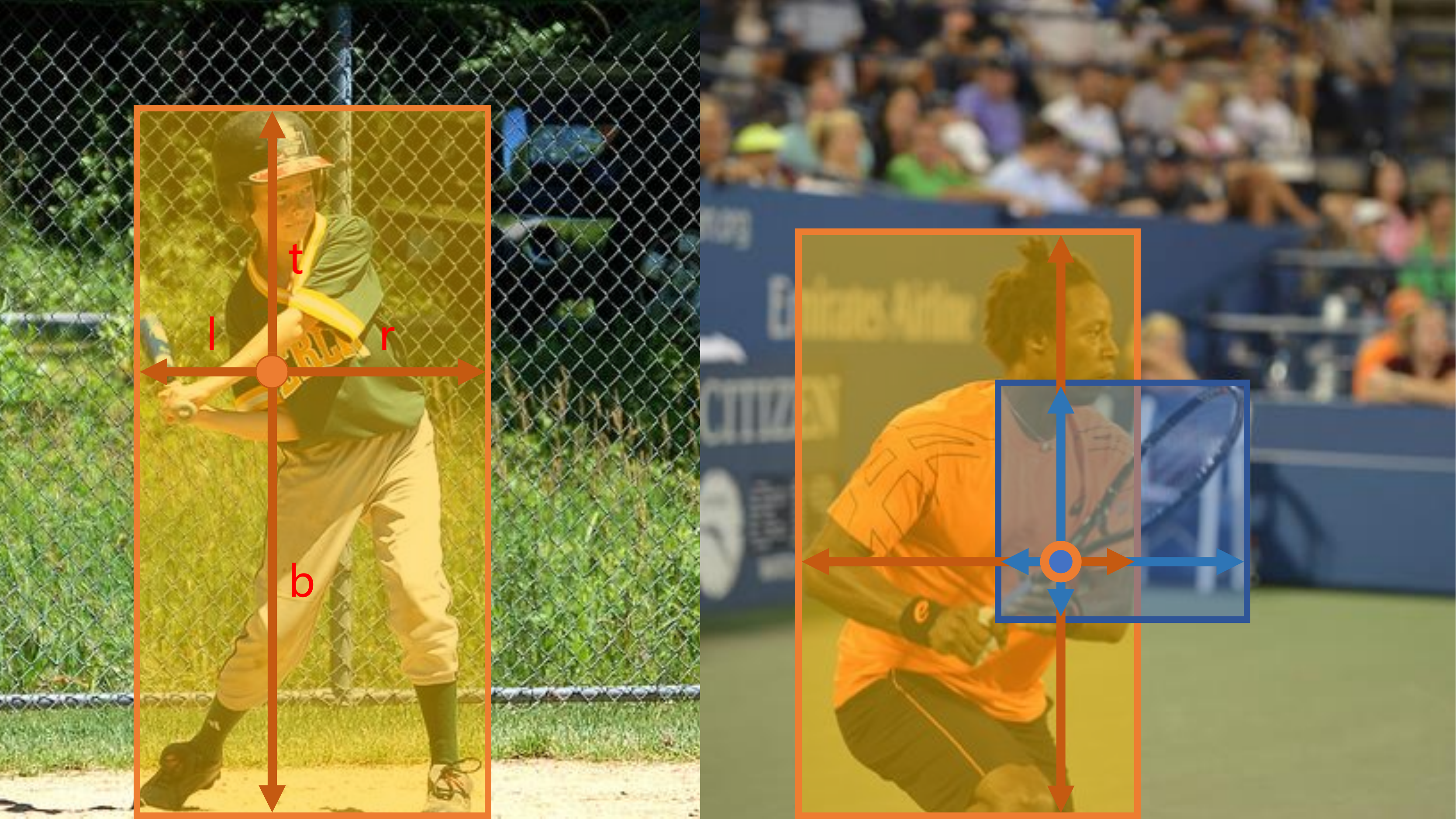}
  \caption{As shown in the left image, \Ours\ works by predicting a 4D vector $(l, t, r, b)$ 
  encoding the location of a bounding box at each foreground pixel (supervised by ground-truth bounding box information
  during training).
  The right plot shows that 
  when a location residing in multiple bounding boxes, it can be ambiguous in terms of which bounding box this location should regress.
}
\label{fig:training_targets}
\end{figure}

Object detection is a fundamental yet 
challenging task
in computer vision, which requires the algorithm to predict a bounding box with a category label for each instance of interest in an image. All current mainstream detectors such as Faster R-CNN \cite{ren2015faster}, SSD \cite{liu2016ssd} and YOLOv2, v3 \cite{redmon2018yolov3} rely on a set of pre-defined anchor boxes and {\it it has long been believed  that 
the use of anchor boxes is  the key to detectors' success.}
Despite their great success, it is important to note that anchor-based detectors suffer some drawbacks:
1) As shown in \cite{lin2017focal, ren2015faster}, 
detection performance is sensitive to 
the sizes, aspect ratios and number of anchor boxes. 
For example, in RetinaNet \cite{lin2017focal}, varying these hyper-parameters affects the performance up to $4\%$ in AP on the COCO benchmark \cite{lin2014microsoft}. As a result, these hyper-parameters need to be carefully tuned  in  anchor-based detectors. 
2) Even with careful design, because
the scales and aspect ratios of anchor boxes are kept fixed, 
detectors encounter difficulties to deal with object candidates with large shape variations, particularly for small objects. The pre-defined anchor boxes also hamper 
the generalization ability of detectors, as they need to be re-designed on new detection tasks with different object sizes or aspect ratios.
3) In order to achieve a high recall rate, an anchor-based detector is required to densely place anchor boxes on the input image (\eg, more than 180K anchor boxes in feature pyramid networks (FPN) \cite{lin2017feature} for an image with its shorter side being 800). Most of these anchor boxes are labelled as negative samples during training. The excessive  number of negative samples aggravates the imbalance between positive and negative samples in training.
4) Anchor boxes also involve complicated computation such as calculating the intersection-over-union (IoU) scores with ground-truth bounding boxes.

Recently, fully convolutional networks (FCNs) \cite{long2015fully} have achieved  
tremendous
success in dense prediction tasks such as semantic segmentation \cite{long2015fully, tian2019decoders, He_2019_CVPR, Liu_2019_CVPR}, depth estimation \cite{Depth2015Liu, Yin2019enforcing}, keypoint detection \cite{ICCV2017Chen} and counting \cite{boominathan2016crowdnet}. As one of high-level vision tasks, object detection might be the only one deviating from the neat fully convolutional per-pixel prediction framework mainly due to the use of anchor boxes.
It is nature to ask a question: {\it Can we solve object detection  
in the neat per-pixel prediction fashion, analogue to FCN for semantic segmentation, for example? 
} Thus those fundamental vision tasks can be unified in (almost) one single framework.
We show that the answer is affirmative. 
Moreover, we demonstrate that, for the first time, the much simpler FCN-based detector achieves even better performance than its anchor-based counterparts.

In the literature, %
some works attempted  to leverage the  FCNs-based framework for object detection such as DenseBox \cite{huang2015densebox}. Specifically, these FCN-based frameworks directly predict a 4D vector plus a class category at each spatial location on a level of feature maps. As shown in  Fig.~\ref{fig:training_targets} (left), the 4D vector depicts the relative offsets from the four sides of a bounding box to the location. These frameworks are  similar to the FCNs for semantic segmentation, except that each location is required to regress a 4D continuous vector.
However, to handle the bounding boxes with different sizes, DenseBox \cite{huang2015densebox} crops and resizes training images to a fixed scale. Thus DenseBox has to perform detection on image pyramids,
which is against FCN's philosophy of computing all convolutions once. 
Besides, 
more significantly, 
these methods %
are mainly used in special domain objection detection such as scene text detection \cite{zhou2017east, he2018end} or face detection \cite{yu2016unitbox, huang2015densebox}, 
since it is believed that these methods 
do not work well when applied to
generic object detection with highly overlapped bounding boxes. As shown in  Fig.~\ref{fig:training_targets} (right), the highly overlapped 
bounding boxes result in an intractable ambiguity:
it is not clear w.r.t.\ which bounding box to regress for the pixels in the overlapped regions.  

In the sequel, we take a closer look at the issue and show that with FPN this ambiguity can be largely eliminated. 
As a result, our method can already obtain comparable detection accuracy with those traditional anchor based detectors.
Furthermore, we %
observe that 
 our method 
may produce a number of low-quality predicted bounding boxes 
at the locations that are far from the center of an target object.   
In order to suppress these low-quality detections,  we introduce a novel ``center-ness''  branch (only one layer) to
predict the deviation of a pixel to the center of its corresponding bounding box, as defined in Eq.~\eqref{eq:centerness}.
This score is then used to down-weight low-quality detected bounding boxes and merge the detection results in NMS. The simple yet effective center-ness branch allows the FCN-based detector to outperform anchor-based counterparts under exactly the same training and testing settings.

This new detection framework enjoys the following advantages.
\begin{itemize}
\itemsep -.05112cm
\item Detection is now unified with many other FCN-solvable tasks such as semantic segmentation, making it easier to re-use ideas from those tasks.
\item Detection becomes proposal free and anchor free, which significantly reduces the number of design parameters. The design parameters typically need heuristic tuning and many tricks are involved in order to achieve good performance. Therefore, our new detection framework makes the detector, particularly its training, \emph{considerably} simpler.
\item By eliminating the anchor boxes, our new detector completely avoids the complicated computation related to anchor boxes such as the IOU computation and matching between the anchor boxes and ground-truth boxes during training, resulting in faster training and testing as well as less training memory footprint than its anchor-based counterpart.
\item Without bells and whistles, we achieve state-of-the-art results among one-stage detectors. We also show that the proposed \Ours\ can be used as a Region Proposal Networks (RPNs) in two-stage detectors and can achieve significantly better performance than its anchor-based RPN counterparts. Given the even better performance of the much simpler anchor-free detector,  {\it we encourage the community to rethink the necessity of anchor boxes in object detection}, which are currently considered as the \emph{de facto} standard for  detection.
\item The proposed detector can be immediately extended to solve other vision tasks with minimal modification, including instance segmentation and  key-point detection. We believe that this new method can be the new baseline for many instance-wise prediction problems.
\end{itemize}

\begin{figure*}[ht!]
\begin{center}
  \includegraphics[width=.85\linewidth]{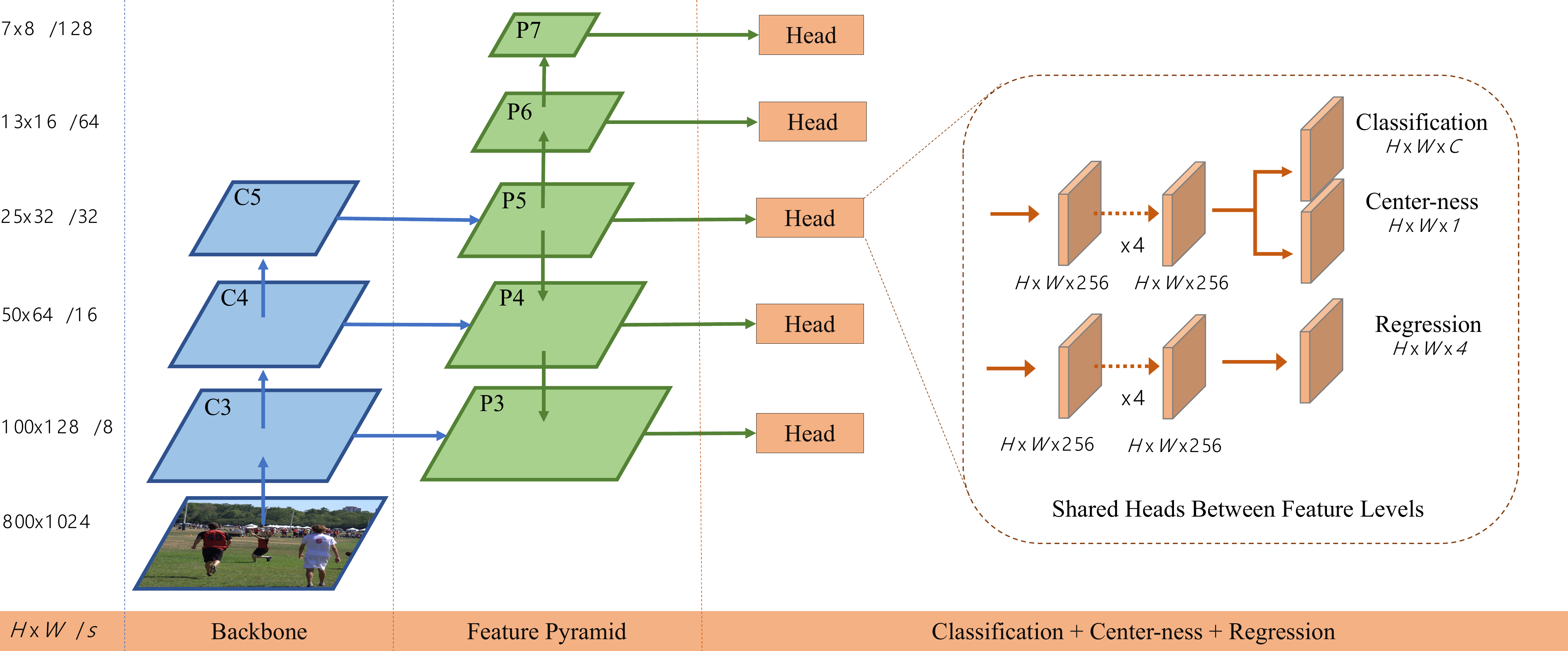}
\end{center}
  \caption{The network architecture of \Ours, where C3, C4, and C5 denote the feature maps of the backbone network and P3 to P7 are the feature levels used for the final prediction. 
  $H \times W$ is the height and width of feature maps. 
  `/$s$' ($s=8, 16, ..., 128$) is the down-sampling ratio of the feature maps at the level to the input image.
  As an example, all the numbers are computed with an $ 800\times 1024$ input.
  }
\label{fig:main_figure}
\end{figure*}

\section{Related Work}
\paragraph{Anchor-based Detectors.} Anchor-based detectors inherit the ideas from traditional sliding-window  and 
proposal based detectors such as Fast R-CNN \cite{girshick2015fast}.
In anchor-based detectors, the anchor boxes can be viewed as pre-defined sliding windows or proposals, which are classified as positive or negative patches, with an extra offsets regression to refine the prediction of bounding box locations. Therefore, the anchor boxes in these detectors may be viewed as \emph{training samples}. Unlike previous detectors like Fast RCNN, which compute image features for each sliding window/proposal repeatedly, 
anchor boxes  make use of the feature maps of CNNs and avoid 
repeated feature computation, speeding up detection process dramatically. The design of anchor boxes are popularized by Faster R-CNN in its RPNs \cite{ren2015faster}, SSD
\cite{liu2016ssd} and YOLOv2 \cite{redmon2017yolo9000},
and has become the %
convention 
in a modern detector.

However, as described above, anchor boxes result in excessively  many hyper-parameters, which typically 
need to be carefully tuned   in order to achieve good performance. Besides the above hyper-parameters describing anchor shapes, the anchor-based detectors also need other hyper-parameters to label each anchor box as a positive, ignored or negative sample.
In previous works, they often employ intersection over union (IOU) between anchor boxes and ground-truth boxes to determine the label of an anchor box (\eg, a positive anchor if its IOU is in $[0.5, 1]$). These hyper-parameters have shown a great impact on the final accuracy, and require heuristic tuning. Meanwhile, these hyper-parameters are specific to detection tasks, making detection tasks deviate from a neat fully convolutional network architectures used in other dense prediction tasks such as semantic segmentation.

\paragraph{Anchor-free Detectors.} The most popular anchor-free detector might be 
YOLOv1 \cite{redmon2016you}. Instead of using anchor boxes, YOLOv1 predicts bounding boxes at points near the center of objects. Only the points near the center are used since they are considered to be able to produce higher-quality detection. However, since only points near the center are used to predict bounding boxes, YOLOv1 suffers from low recall as mentioned in YOLOv2 \cite{redmon2017yolo9000}. As a result, YOLOv2 \cite{redmon2017yolo9000} employs anchor boxes as well. Compared to YOLOv1, \Ours\ takes advantages of all points in a ground truth bounding box to predict the bounding boxes and the low-quality detected bounding boxes are suppressed by the proposed ``center-ness" branch.  As a result, \Ours\ is able to provide comparable recall with anchor-based detectors as shown in our experiments.

CornerNet \cite{law2018cornernet} is a recently proposed one-stage anchor-free detector, which detects a pair of corners of a bounding box and groups them to form the final detected bounding box. %
CornerNet requires much more complicated post-processing to group the pairs of corners belonging to the same instance. An extra distance metric is learned for the purpose of grouping.

Another family of anchor-free detectors such as \cite{yu2016unitbox} are based on DenseBox \cite{huang2015densebox}. %
 The family of detectors have been considered  unsuitable for generic object detection due to 
 difficulty in handling overlapping bounding boxes and  the recall being relatively low.
 In this work, we show that both problems can be largely alleviated  with multi-level FPN prediction. Moreover, we also show together with our proposed center-ness branch, the much simpler detector can achieve even better detection performance than its anchor-based counterparts.

\section{Our Approach}
In this section, we first reformulate object detection in a per-pixel prediction fashion. Next, we show that how we make use of multi-level prediction to improve the recall and resolve the ambiguity resulted from overlapped bounding boxes. Finally, we present our proposed ``center-ness" branch, which helps suppress the low-quality detected bounding boxes and improves the overall performance by a large margin.

\subsection{Fully Convolutional One-Stage Object Detector%
}
Let $F_i \in \R^{H \times W \times C}$ be the feature maps at layer $i$ of a backbone CNN and $s$ be the total stride until the layer. The ground-truth bounding boxes for an input image are defined as $\{B_i\}$, where $B_i = (x^{(i)}_0, y^{(i)}_0, x^{(i)}_1 y^{(i)}_1, c^{(i)}) \in \R^4 \times \{1, 2\ ...\ C\}$. 
Here 
$(x^{(i)}_0, y^{(i)}_0)$ and $(x^{(i)}_1 y^{(i)}_1)$ %
denote the coordinates of the left-top 
and right-bottom corners of the bounding box.
$c^{(i)}$ is 
the class that the object in the bounding box belongs to. %
$C$ is the number of classes, which is $80$ for MS-COCO dataset.

For each location $(x, y)$ on the feature map $F_i$, we can map it back onto the input image as $(\floor{\frac{s}{2}} + xs, \floor*{\frac{s}{2}} + ys)$, which is near the center of the receptive field of the location $(x, y)$. Different from anchor-based detectors, which consider the location on the input image as the center of (multiple) anchor boxes and 
regress the target bounding box with these anchor boxes as references, we directly %
regress the target bounding box at the location. In other words, our detector directly views locations as \emph{training samples} instead of anchor boxes in anchor-based detectors, which is the same as FCNs for semantic segmentation \cite{long2015fully}.

Specifically, location $(x, y)$ is considered as a positive sample if it falls into any ground-truth box and the class label $c^*$ of the location is the class label of the ground-truth box.
Otherwise it is a negative sample and $c^* = 0$ (background class). Besides the label for classification, we also have a 4D real vector $\vt^* = (l^*, t^*, r^*, b^*)$ 
being the regression targets for the location. Here $l^*$, $t^*$, $r^*$ and $b^*$ are the distances from the location to the four sides of the bounding box,
as shown in Fig.~\ref{fig:training_targets} (left). If a location falls into multiple bounding boxes, it is considered as an \emph{ambiguous sample}. We simply choose the bounding box with minimal area as its regression target. In the next section, we will show that with multi-level prediction, the number of ambiguous samples can be reduced significantly and thus they hardly affect the detection performance. Formally, if location $(x, y)$ is associated to a bounding box $B_i$, the training regression targets for the location can be formulated as,
\begin{equation} \label{regression_targets}
\begin{aligned}
    l^* = x - x^{(i)}_0,~~t^* = y - y^{(i)}_0, \\
    r^* = x^{(i)}_1 - x,~~b^* = y^{(i)}_1 - y.
\end{aligned}
\end{equation}
{\it It is worth noting that \Ours\ can leverage as many foreground samples as possible to train the regressor.} It is different from anchor-based detectors, which only consider the anchor boxes with a highly enough IOU with ground-truth boxes as positive samples. We argue that it may be one of the reasons that \Ours\ outperforms its anchor-based counterparts.

\paragraph{Network Outputs.} Corresponding to the training targets, the final layer of our networks predicts an 80D vector $\vp$ 
of classification labels
and a 4D vector  $\vt =(l, t, r, b)$ 
bounding box coordinates. 
Following \cite{lin2017focal}, instead of training a multi-class classifier, 
we train $ C $ binary classifiers. 
Similar to \cite{lin2017focal}, we add four convolutional layers after the feature maps of the backbone networks respectively for classification and regression branches. 
Moreover, since the regression targets are always positive, we employ $\exp(x)$ to map any real number to $(0, \infty)$ on the top of the regression branch. \emph{It is worth noting that \Ours\ has $9\times$ fewer network output variables  than the popular anchor-based detectors \cite{lin2017focal, ren2015faster} with 9 anchor boxes per location.}

\def\cls{ {\rm cls} }
\def\reg{ {\rm reg} }
\def\pos{ {\rm pos} }

\paragraph{Loss Function.} We define our training loss function as follows:
\begin{align} 
\label{loss_function}
    L(\{\vp_{x, y}\}, \{\vt_{x, y}\}) &= \frac{1}{N_{\pos}} \sum_{x, y}{L_{\cls}(\vp_{x, y}, c^*_{x, y})} \nonumber \\
    & + \frac{\lambda}{N_{\pos}}\sum_{{x, y}}{\mathbbm{1}_{\{c^*_{x, y} > 0\}}L_{\reg}(\vt_{x, y}, \vt^*_{x, y})},
\end{align}
where $L_{\cls}$ is focal loss as in \cite{lin2017focal} and $L_{\reg}$ is the IOU loss as in UnitBox \cite{yu2016unitbox}.  $N_{\pos}$ denotes the number of positive samples and $\lambda$ being $1$ in this paper is the balance weight for $L_{\reg}$. The summation is calculated  over all locations on the feature maps $F_i$. $\mathbbm{1}_{\{c^*_i > 0\}}$ is the indicator function, being $1$ if $c^*_i > 0$ and $0$ otherwise.

\paragraph{Inference.} The inference of \Ours\ is straightforward. Given an input images, we forward it through the network and obtain the classification 
scores 
$\vp_{x, y}$ 
and the regression prediction $\vt_{x, y}$ for each location on the feature maps $F_i$. Following \cite{lin2017focal}, we choose the location with $p_{x, y} > 0.05$ as positive samples and invert Eq.~(\ref{regression_targets}) to obtain the predicted bounding boxes.

\subsection{Multi-level Prediction with FPN for \Ours}
Here we show that how two possible issues of the proposed \Ours\ can be resolved with multi-level prediction with FPN \cite{lin2017feature}.
1) The large stride (\eg, 16$\times$) of the final feature maps in a 
CNN  %
can result in a relatively low {\it best possible recall (BPR)\footnote{Upper bound of the recall rate that a detector can achieve.}}.
For anchor based detectors, low recall rates due to the large stride can be compensated to some extent by lowering  the required IOU scores for positive anchor boxes. 
For \Ours, at the first glance 
one may think that the BPR can be much lower than anchor-based detectors because   
it is impossible to recall an object %
which no location on the final feature maps  
encodes    
due to a large stride.  %
Here,  
we empirically show that even with a large stride, FCN-based \Ours\ is still able to produce a good BPR, and it  can even better than the BPR of the anchor-based detector RetinaNet \cite{lin2017focal} in the official implementation  Detectron \cite{Detectron2018}
 (refer to Table \ref{table:recall}).
 Therefore, the BPR is actually not a problem of \Ours. Moreover, with multi-level  FPN prediction  \cite{lin2017feature}, the BPR can be improved further to match 
the best BPR the anchor-based RetinaNet can achieve. 2) Overlaps  in ground-truth boxes can cause 
intractable ambiguity %
, {\it i.e.}, which bounding box should a location in the overlap regress? 
This ambiguity results in degraded performance of  FCN-based detectors. In this work, we show that the ambiguity can be greatly resolved with multi-level prediction, and the FCN-based detector can obtain {\it on par}, sometimes even better, performance compared with anchor-based ones.

Following FPN \cite{lin2017feature}, we detect different sizes of objects on different levels of feature maps. Specifically, we make use of five levels of feature maps defined as $\{P_3, P_4, P_5, P_6, P_7\}$. $P_3$, $P_4$ and $P_5$ are produced by the backbone CNNs' feature maps $C_3$, $C_4$ and $C_5$ followed by a $1 \times 1$ convolutional layer with the top-down connections in \cite{lin2017feature}, as shown in Fig.~\ref{fig:main_figure}. $P_6$ and $P_7$ are produced by applying one convolutional layer with the stride being 2 on $P_5$ and $P_6$, respectively. As a result, the feature levels $P_3$, $P_4$, $P_5$, $P_6$ and $P_7$ have strides 8, 16, 32, 64 and 128, respectively.

Unlike anchor-based detectors, which assign anchor boxes with different sizes to different feature levels, we directly limit the range of bounding box regression for each level.
More specifically, we firstly compute the regression targets $l^*$, $t^*$, $r*$ and $b^*$ for each location on all feature levels. Next, if a location satisfies
$\max(l^*, t^*, r^*, b^*) > m_i$ or $\max(l^*$, $ t^*, r^*, b^*) < m_{i-1}$, it is set as a negative sample and is thus not required to regress a bounding box anymore.
Here $m_i$ is the maximum distance that feature level $i$ needs to regress.
In this work, $m_2$, $m_3$, $m_4$, $m_5$, $m_6$ and $m_7$ are set as $0$, $64$, $128$, $256$, $512$ and $\infty$, respectively. Since objects with different sizes are assigned to different feature levels and most overlapping happens between objects with %
considerably 
different sizes. If a location, even with multi-level prediction used, is still assigned to more than one ground-truth boxes, we simply choose the ground-truth box with minimal area as its target. As shown in our experiments, the multi-level prediction can %
largely alleviate 
the aforementioned ambiguity and improve the FCN-based detector to the same level of anchor-based ones.

Finally, following \cite{lin2017feature, lin2017focal}, we share the heads between different feature levels,  not only making  the detector parameter-efficient but also improving the detection performance. %
However, we observe that different feature levels are required to regress different size range (\eg, the size range is $[0, 64]$ for $P_3$ and $[64, 128]$ for $P_4$), and therefore it is not reasonable to make use of identical heads for different feature levels. As a result, instead of using the standard $\exp(x)$, we make use of $\exp(s_ix)$ with a trainable scalar $s_i$ to automatically adjust the base of the exponential function for feature level $P_i$, which slightly improves the detection performance.

\subsection{Center-ness for \Ours}
After using multi-level prediction in \Ours, there is still a performance gap between \Ours\ and anchor-based detectors. We observed that it is due to a lot of low-quality predicted bounding boxes produced by locations far away from the center of an object.

We propose a simple yet effective strategy to suppress these low-quality detected bounding boxes without introducing any hyper-parameters. Specifically, we add {\it a  single-layer branch}, in parallel with the classification branch (as shown in Fig.~\ref{fig:main_figure}) to predict the ``center-ness" of a location\footnote{After the initial submission, it has been shown that the AP on MS-COCO can be improved if the center-ness is parallel with the regression branch instead of the classification branch. However, unless specified, we still use the configuration in Fig. \ref{fig:main_figure}.}. The center-ness depicts the normalized distance from the location to the center of the object that the location is responsible for, as shown Fig.~\ref{fig:centerness}. Given the regression targets $l^*$, $t^*$, $r^*$ and $b^*$ for a location, the center-ness target is defined as,
\begin{equation}
\label{eq:centerness}
    \centerness^* = \sqrt{\frac{ \min(l^*, r^*)}{ \max(l^*, r^*)} \times \frac{ \min(t^*, b^*)}{ \max(t^*, b^*)}}.
\end{equation}
We employ $\rm sqrt$ here to slow down the decay of the center-ness. The center-ness ranges from $0$ to $1$ and is thus trained with binary cross entropy (BCE) loss. The loss is added to the loss function Eq.~(\ref{loss_function}). When testing, the final score (used for ranking the detected bounding boxes) is computed by multiplying the predicted center-ness with the corresponding classification score. Thus the center-ness can down-weight the scores of bounding boxes far from the center of an object. As a result, with high probability, these low-quality bounding boxes might be filtered out by the final non-maximum suppression (NMS) process, improving the detection performance \emph{remarkably}.

\begin{figure}[t]
\centering
\includegraphics[width=.22730524\textwidth,trim={0 0 0 0},clip]{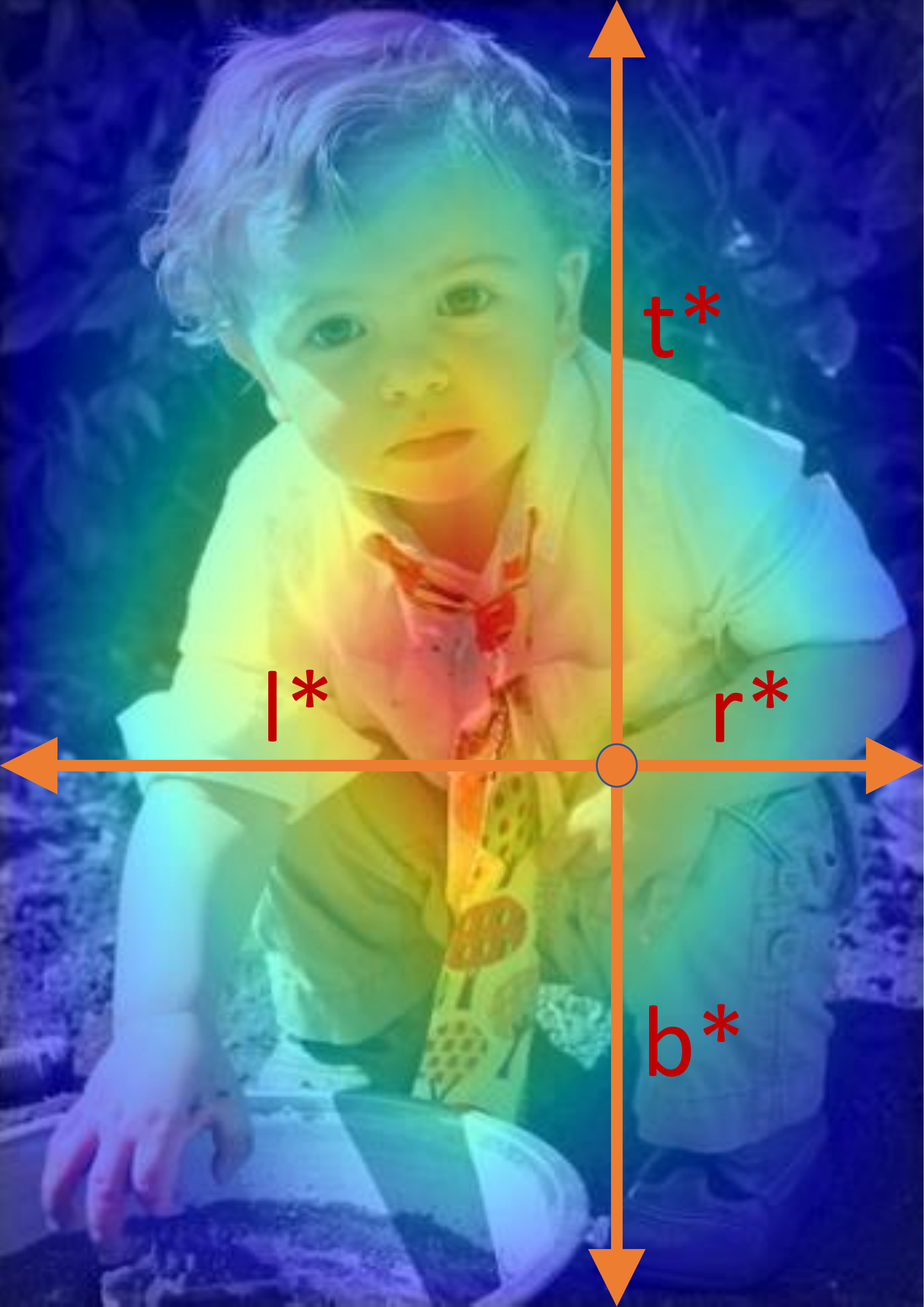}
\caption{\textbf{Center-ness.}
Red, blue, and other colors denote 1, 0 and the values between them, respectively. Center-ness is computed by Eq. (\ref{eq:centerness}) and decays from 1 to 0 as the location deviates from the center of the object. When testing, the center-ness predicted by the network is multiplied with the classification score thus can down-weight the low-quality bounding boxes predicted by a location far from the center of an object.}
\label{fig:centerness}
%
\end{figure}

An alternative of the center-ness is to make use of only the central portion of ground-truth bounding box as positive samples with the price of one extra hyper-parameter, as shown in works \cite{huang2015densebox, zhou2017east}. After our submission, it has been shown in \cite{fcosplus} that the combination of both methods can achieve a much better performance. The experimental results can be found in Table~\ref{table:ours_vs_anchor_based_detector}.

\section{Experiments}
Our experiments are conducted on the large-scale detection benchmark COCO \cite{lin2014microsoft}. Following the common practice \cite{lin2017focal, lin2017feature, ren2015faster}, we use the COCO  \texttt{trainval35k} split (115K images) for training and \texttt{minival} split (5K images) as validation for our ablation study. We report our main results on the \texttt{test\_dev} split (20K images) by uploading our detection results to the evaluation  server.

\paragraph{Training Details.} Unless specified, ResNet-50 \cite{he2016deep} is used as our backbone networks and the same hyper-parameters with RetinaNet \cite{lin2017focal} are used. Specifically, our network is trained with stochastic gradient descent (SGD) for 90K iterations with the initial learning rate being 0.01 and a mini-batch of 16 images. The learning rate is reduced by a factor of 10 at iteration 60K and 80K, respectively. Weight decay and momentum are set as 0.0001 and 0.9, respectively. We initialize our backbone networks with the weights pre-trained on ImageNet \cite{deng2009imagenet}. For the newly added layers, we initialize them as in \cite{lin2017focal}. Unless specified, the input images are resized to have their shorter side being 800 and their longer side less or equal to 1333.

\paragraph{Inference Details.} We firstly forward the input image through the network and obtain the predicted bounding boxes with a predicted class. Unless specified, the following post-processing is exactly the same with RetinaNet \cite{lin2017focal} and we directly make use of the same post-processing hyper-parameters of RetinaNet. 
We use the same sizes of input images as in training.
We hypothesize  that  the performance of our detector may be 
improved further if we carefully tune the hyper-parameters.

\subsection{Ablation Study}
\subsubsection{Multi-level Prediction with FPN}
\begin{table}[!t]
\centering 
\small
\begin{tabular}{ l|c|c|c }
Method & w/ FPN & Low-quality matches & BPR (\%) \\
\Xhline{2\arrayrulewidth}
RetinaNet & \checkmark & None & 86.82 \\
RetinaNet & \checkmark & $\geq 0.4$ & 90.92 \\
RetinaNet & \checkmark & All & \textbf{99.23} \\
\hline
\Ours &  & - & 95.55 \\
\Ours & \checkmark & - & 98.40 \\
\end{tabular}
\caption{The BPR for anchor-based RetinaNet under a variety of matching rules and the BPR for FCN-based \Ours. FCN-based \Ours\ has very similar recall to the best anchor-based one and has much higher recall than the official implementation in Detectron \cite{Detectron2018}, where only low-quality matches with IOU $\geq0.4$ are considered.}
\label{table:recall}
\end{table}

\begin{table}[!t]
\centering 
\small
\begin{tabular}{ c|c|c }
w/ FPN & Amb. samples (\%) & Amb. samples (diff.) (\%) \\
\Xhline{2\arrayrulewidth}
 & 23.16 & 17.84 \\
\checkmark & \textbf{7.14} & \textbf{3.75} \\
\end{tabular}
\caption{
Amb. samples denotes the ratio of the ambiguous samples to all positive samples. Amb. samples (diff.) is similar but excludes those ambiguous samples in the overlapped regions but belonging to the same category as the kind of ambiguity does not matter when inferring. We can see that with FPN, this percentage of ambiguous samples is small (3.75\%).
}
\label{table:overlapping}
\vspace{-0.5cm}
\end{table}

\begin{table*}[!tbh]
\centering 
\small
\begin{tabular}{ l | c |c|c|c| c c | c c c | c c c}
Method & $C_5$/$P_5$ & w/ GN & nms thr. & AP & AP$_{50}$ & AP$_{75}$ & AP$_{S}$ & AP$_{M}$ & AP$_{L}$ & AR$_{1}$ & AR$_{10}$ & AR$_{100}$ \\
\Xhline{2\arrayrulewidth}
RetinaNet & $C_5$ & & .50 & 35.9 & 56.0 & 38.2 & 20.0 & 39.8 & 47.4 & 31.0 & 49.4 & 52.5 \\
\Ours\ & $C_5$ & & .50 & 36.3 & 54.8 & 38.7 & 20.5 & 39.8 & 47.8 & 31.5 & 50.6 & 53.5 \\
\Ours\ & $P_5$ & & .50 & 36.4 & 54.9 & 38.8 & 19.7 & 39.7 & 48.8 & 31.4 & 50.6 & 53.4 \\
\hline
\Ours\ & $P_5$ & & .60 & 36.5 & 54.5 & 39.2 & 19.8 & 40.0 & 48.9 & 31.3 & 51.2 & 54.5 \\
\Ours\ & $P_5$ & \checkmark & .60 & 37.1 & 55.9 & 39.8 & 21.3 & 41.0 & 47.8 & 31.4 & 51.4 & 54.9 \\
\hline
\textbf{Improvements} & & & & & & & & & \\
+ ctr.\ on reg.\ & $P_5$ & \checkmark & .60 & 37.4 & 56.1 & 40.3 & 21.8 & 41.2 & 48.8 & 31.5 & 51.7 & 55.2 \\
+ ctr.\  sampling \cite{fcosplus} & $P_5$ & \checkmark & .60 & 38.1 & 56.7 & \textbf{41.4} & \textbf{22.6} & 41.6 & \textbf{50.4} & 32.1 & 52.8 & 56.3 \\
+ GIoU \cite{fcosplus} & $P_5$ & \checkmark & .60 & 38.3 & 57.1 & 41.0 & 21.9 & 42.4 & 49.5 & 32.0 & 52.9 & 56.5 \\
+ Normalization & $P_5$ & \checkmark & .60 & \textbf{38.6} & \textbf{57.4} & \textbf{41.4} & 22.3 & \textbf{42.5} & 49.8 & \textbf{32.3} & \textbf{53.4} & \textbf{57.1} \\
\end{tabular}
\caption{\Ours\ vs.\ RetinaNet on the \texttt{minival} split with ResNet-50-FPN as the backbone. Directly using the training and testing settings of RetinaNet, our anchor-free \Ours\ achieves even better performance than anchor-based RetinaNet both in AP and AR. With Group Normalization (GN) in heads and NMS threshold being 0.6, \Ours\ can achieve $37.1$ in AP. After our submission, some almost cost-free improvements have been made for \Ours\ and the performance has been improved by a large margin, as shown by the rows %
below
``\textbf{Improvements}". ``ctr. on reg.": moving the center-ness branch to the regression branch. ``ctr. sampling": only sampling the central portion of ground-truth boxes as positive samples. ``GIoU": penalizing the union area over the circumscribed rectangle's area in IoU Loss. ``Normalization": normalizing the regression targets in Eq.~(\ref{regression_targets}) with the strides of FPN levels. Refer to our code for details.}
\label{table:ours_vs_anchor_based_detector}
\vspace{-0.7cm}
\end{table*}

As mentioned before, the major concerns of an FCN-based detector are \emph{low recall} rates and \emph{ambiguous samples} resulted from overlapping in ground-truth bounding boxes. In the section, we show that both issues can be largely resolved with multi-level prediction.

\paragraph{Best Possible Recalls.} The first concern about the FCN-based detector is that it might not provide a good {best possible recall (BPR)}. In the section, we show that the concern is not necessary. Here BPR is 
defined as the ratio of the number of ground-truth boxes a detector can recall at the most divided by all ground-truth boxes. A ground-truth box is considered being recalled if the box is assigned to at least one sample (\ie, a location in \Ours\ or an anchor box in anchor-based detectors) during training.  As shown in Table~\ref{table:recall}, only with feature level $P_4$ with stride being 16 (\ie, no FPN), \Ours\ can already obtain a BPR of $95.55\%$.
The BPR is much higher than the BPR of $90.92\%$ of the anchor-based detector RetinaNet in the official implementation Detectron, where only the low-quality matches with IOU $\geq0.4$ are used. With the help of FPN, \Ours\ can achieve a BPR of $98.40\%$, which is very close to the best BPR that the anchor-based detector can achieve by using all low-quality matches. Due to the fact that the best recall of current detectors are much lower than $90\%$, the small BPR gap (less than $1\%$) between \Ours\ and the anchor-based detector will not actually affect the performance of detector. It is also confirmed in Table~\ref{table:ours_vs_anchor_based_detector}, where \Ours\ achieves even better AR than its anchor-based counterparts under the same training and testing settings. Therefore, the concern about low BPR may not be necessary.

\paragraph{Ambiguous Samples.} Another concern about the FCN-based detector is that it may have a large number of \emph{ambiguous samples} due to the overlapping in ground-truth bounding boxes, as shown in Fig.~\ref{fig:training_targets} (right). In Table~\ref{table:overlapping}, we show the ratios of the ambiguous samples to all positive samples on \texttt{minival} split. As shown in the table, there are indeed a large amount of ambiguous samples ($23.16\%$) if FPN is not used and only feature level $P_4$ is used. However, with FPN, the ratio can be significantly reduced to only $7.14\%$ since most of overlapped objects are assigned to different feature levels. Moreover, we argue that the ambiguous samples resulted from overlapping between objects of the same category do not matter. For instance, if object A and B with the same class have overlap, no matter which object the locations in the overlap predict, the prediction is correct because it is always matched with the same category. The missed object can be predicted by the locations only belonging to it. Therefore, we only count the ambiguous samples in overlap between bounding boxes with different categories. As shown in Table~\ref{table:overlapping}, the multi-level prediction reduces the ratio of ambiguous samples from $17.84\%$ to $3.75\%$. In order to further show that the overlapping in ground truth boxes is not a problem of our FCN-based \Ours, we count that when inferring how many detected bounding boxes come from the ambiguous locations. We found that only $2.3\%$ detected bounding boxes are produced by the ambiguous locations. By further only considering the overlap between different categories, the ratio is reduced to $1.5\%$. Note that it does not imply that there are $1.5\%$ locations where \Ours\ cannot work. As mentioned before, these locations are associated with the ground-truth boxes with minimal area. Therefore, these locations only take the risk of missing some larger objects. As shown in the following experiments, they do not make our \Ours\ inferior to anchor-based detectors.

\subsubsection{With or Without Center-ness}
As mentioned before, we propose ``center-ness" to suppress the low-quality detected bounding boxes produced by the locations far from the center of an object. As shown in Table~\ref{table:center-ness}, the center-ness branch can boost AP from $33.5\%$ to $37.1\%$, making anchor-free \Ours\ outperform anchor-based RetinaNet ($35.9\%$). Note that anchor-based RetinaNet employs two IoU thresholds to label anchor boxes as positive/negative samples, which can also help to suppress the low-quality predictions. The proposed center-ness can eliminate the two hyper-parameters. However, after our initial submission, it has shown that using both center-ness and the thresholds can result in a better performance, as shown by the row ``+ ctr. sampling" in Table \ref{table:ours_vs_anchor_based_detector}. One may note that center-ness can also be computed with the predicted regression vector without introducing the extra center-ness branch. However, as shown in Table~\ref{table:center-ness}, the center-ness computed from the regression vector cannot improve the performance and thus the separate center-ness is necessary.

\subsubsection{\Ours\ vs.\ Anchor-based Detectors}

\begin{table}
\centering 
\small
\begin{tabular}{ l|c|c c|c c c }
 & AP & AP$_{50}$ & AP$_{75}$ & AP$_{S}$ & AP$_{M}$ & AP$_{L}$ \\
\Xhline{2\arrayrulewidth}
None & 33.5 & 52.6 & 35.2 & 20.8 & 38.5 & 42.6 \\
center-ness$^\dag$ & 33.5 & 52.4 & 35.1 & 20.8 & 37.8 & 42.8 \\
center-ness & \textbf{37.1} & \textbf{55.9} & \textbf{39.8} & \textbf{21.3} & \textbf{41.0} & \textbf{47.8} \\
\end{tabular}
\caption{Ablation study for the proposed center-ness branch on \texttt{minival} split. ``None" denotes that  no center-ness is used.  ``center-ness$^\dag$" denotes that using the center-ness computed from the predicted regression vector. ``center-ness" is that using center-ness predicted from the proposed center-ness branch. The center-ness branch improves the detection performance under all metrics.}
\label{table:center-ness}
\vspace{-0.7cm}
\end{table}

\begin{table*}[!htb]
\centering 
\small
\begin{tabular}{ l | l |c c c | c c c }
Method & Backbone & AP & AP$_{50}$ & AP$_{75}$ & AP$_{S}$ & AP$_{M}$ & AP$_{L}$ \\
\Xhline{2\arrayrulewidth}
\hspace{-0.25cm}\textrm{Two-stage methods:} &&&&&&& \\
Faster R-CNN w/ FPN \cite{lin2017feature} & ResNet-101-FPN & 36.2 & 59.1 & 39.0 & 18.2 & 39.0 & 48.2 \\
Faster R-CNN by G-RMI \cite{huang2017speed} & Inception-ResNet-v2 \cite{szegedy2017inception} & 34.7 & 55.5 & 36.7 & 13.5 & 38.1 & 52.0 \\
Faster R-CNN w/ TDM \cite{shrivastava2016beyond} & Inception-ResNet-v2-TDM & 36.8 & 57.7 & 39.2 & 16.2 & 39.8 & 52.1 \\
\hline
\hspace{-0.25cm}\textrm{One-stage methods:} &&&&&&& \\
YOLOv2 \cite{redmon2017yolo9000} & DarkNet-19 \cite{redmon2017yolo9000} & 21.6 & 44.0 & 19.2 & 5.0 & 22.4 & 35.5 \\
SSD513 \cite{liu2016ssd} & ResNet-101-SSD & 31.2 & 50.4 & 33.3 & 10.2 & 34.5 & 49.8 \\
DSSD513 \cite{fu2017dssd} & ResNet-101-DSSD & 33.2 & 53.3 & 35.2 & 13.0 & 35.4 & 51.1 \\
RetinaNet \cite{lin2017focal} & ResNet-101-FPN & 39.1 & 59.1 & 42.3 & 21.8 & 42.7 & 50.2 \\
CornerNet \cite{law2018cornernet} & Hourglass-104 & 40.5 & 56.5 & 43.1 & 19.4 & 42.7 & 53.9 \\
FSAF \cite{zhu2019feature} & ResNeXt-64x4d-101-FPN & 42.9 & 63.8 & 46.3 & 26.6 & 46.2 & 52.7 \\
\hline
\Ours & ResNet-101-FPN & 41.5 & 60.7 & 45.0 & 24.4 & 44.8 & 51.6 \\
\Ours & HRNet-W32-5l \cite{DBLP:journals/corr/abs-1902-09212} & 42.0 & 60.4 & 45.3 & 25.4 & 45.0 & 51.0 \\
\Ours & ResNeXt-32x8d-101-FPN & 42.7 & 62.2 & 46.1 & 26.0 & 45.6 & 52.6 \\
\Ours & ResNeXt-64x4d-101-FPN & 43.2 & 62.8 & 46.6 & 26.5 & 46.2 & 53.3 \\
\hline
FCOS w/ improvements & ResNeXt-64x4d-101-FPN & \textbf{44.7} & \textbf{64.1} & \textbf{48.4} & \textbf{27.6} & \textbf{47.5} & \textbf{55.6} \\
\end{tabular}
\caption{\Ours\ vs.\ other state-of-the-art two-stage or one-stage detectors (\emph{single-model and single-scale results}). \Ours\ outperforms the anchor-based counterpart RetinaNet by $2.4\%$ in AP with the same backbone. \Ours\ also outperforms  the recent anchor-free one-stage detector CornerNet with much less design complexity. Refer to Table \ref{table:ours_vs_anchor_based_detector} for details of ``improvements''.}

\label{table:ours_vs_stoa_detectors}
\vspace{-0.5cm}
\end{table*}

\begin{table}[!htp]
\centering 
\small
\begin{tabular}{  l |c|c|c }
Method & \# samples & AR$^{100}$ & AR$^{1k}$ \\
\Xhline{2\arrayrulewidth}
RPN w/ FPN \& GN (ReImpl.) & $\sim$200K & 44.7 & 56.9 \\
\Ours\ w/ GN w/o center-ness & $\sim$66K & 48.0 & 59.3 \\
\Ours\ w/ GN & $\sim$66K & \textbf{52.8} & \textbf{60.3} \\
\end{tabular}
\caption{\Ours\ as Region Proposal Networks vs.\ 
RPNs with FPN. ResNet-50 is used as the backbone. \Ours\ improves AR$^{100}$ and AR$^{1k}$ by $8.1\%$ and $3.4\%$, respectively. GN: Group Normalization.}
\label{table:rpn_fpn}
\vspace{-0.7cm}
\end{table}

The aforementioned \Ours\ has two minor differences from the standard RetinaNet. 1) We use Group Normalization (GN) \cite{wu2018group} in the newly added convolutional layers except for the last prediction layers, which makes our training more stable. 2) We use $P_5$ to produce the $P_6$ and $P_7$ instead of $C_5$ in the standard RetinaNet. We observe that using $P_5$ can improve the performance slightly.

To show that our \Ours\ can serve as an simple and strong alternative of anchor-based detectors, and for a fair comparison, 
we remove GN (the gradients are clipped to prevent them from exploding) and use $C_5$ in our detector. As shown in Table~\ref{table:ours_vs_anchor_based_detector}, with exactly the same settings, our \Ours\ still compares favorably with the anchor-based detector ($36.3\%$ vs $35.9\%$). Moreover, it is worth to note that we directly use all hyper-parameters (\eg, learning rate, the NMS threshold and etc.) from RetinaNet, which have been optimized for the anchor-based detector. We argue that the performance of \Ours\ can be improved further if the hyper-parameters are tuned for it.

It is worth noting that with some almost cost-free improvements, as shown in Table~\ref{table:ours_vs_anchor_based_detector}, the performance of our anchor-free detector can be improved by a large margin. Given the superior performance and the merits of the anchor-free detector (\eg, much simpler and fewer hyper-parameters than anchor-based detectors), we encourage the community to rethink the necessity of anchor boxes in object detection.

\subsection{Comparison with State-of-the-art Detectors}
We compare \Ours\ with other state-of-the-art object detectors on \texttt{test-dev} split of MS-COCO benchmark. For these experiments, we randomly scale the shorter side of images in the range from 640 to 800 during the training and double the number of iterations to 180K (with the learning rate change points scaled proportionally). Other settings are exactly the same as the model with AP $37.1\%$ in Table~\ref{table:ours_vs_anchor_based_detector}. As shown in Table \ref{table:ours_vs_stoa_detectors}, with ResNet-101-FPN, our \Ours\ outperforms the RetinaNet with the same backbone ResNet-101-FPN by $2.4\%$ in AP. To our 
knowledge, it is the first time that an anchor-free detector, without any bells and whistles outperforms anchor-based detectors by a large margin. \Ours\ also outperforms other classical two-stage anchor-based detectors such as Faster R-CNN by a large margin. With ResNeXt-64x4d-101-FPN \cite{xie2017aggregated} as the backbone, \Ours achieves $43.2\%$ in AP. It outperforms the recent state-of-the-art anchor-free detector CornerNet \cite{law2018cornernet} by a large margin while being much simpler. Note that CornerNet requires to group corners with embedding vectors, which needs special design for the detector. Thus, we argue that \Ours\ is more likely to serve as a strong and simple alternative to current mainstream anchor-based detectors. Moreover, \Ours\ with the improvements in Table \ref{table:ours_vs_anchor_based_detector} achieves $44.7\%$ in AP with single-model and single scale testing, which surpasses previous detectors by a large margin.

\section{Extensions on Region Proposal Networks}
So far we have shown that in a one-stage detector, our \Ours\ can achieve even better performance than anchor-based counterparts. Intuitively, \Ours\ should be also able to replace the anchor boxes in Region Proposal Networks (RPNs) with FPN \cite{lin2017feature} in the two-stage detector Faster R-CNN. Here, we confirm that by experiments.

Compared to RPNs with FPN \cite{lin2017feature}, we replace anchor boxes with the method in \Ours. Moreover, we add GN into the layers in FPN heads, which can make our training more stable. All other settings are exactly the same with RPNs with FPN in the official code \cite{Detectron2018}. As shown in Table \ref{table:rpn_fpn}, even without the proposed center-ness branch, our \Ours\ already improves both AR$^{100}$ and AR$^{1k}$ significantly. With the proposed center-ness branch, \Ours\ further boosts AR$^{100}$ and AR$^{1k}$ respectively to $52.8\%$ and $60.3\%$, which are $18\%$ relative improvement for AR$^{100}$ and $3.4\%$ absolute improvement for AR$^{1k}$ over the RPNs with FPN.

\section{Conclusion}
We have proposed an anchor-free and proposal-free one-stage detector \Ours. As shown in experiments, \Ours\ compares favourably against the popular anchor-based one-stage detectors, including RetinaNet, YOLO and SSD, but with much less design complexity. \Ours\ completely avoids all computation and hyper-parameters related to anchor boxes and solves the object detection in a per-pixel prediction fashion, similar to other dense prediction tasks such as semantic segmentation.  \Ours\ also achieves state-of-the-art performance among one-stage detectors. We also show that \Ours\ can be used as RPNs in the two-stage detector Faster R-CNN and outperforms the its RPNs by a large margin. Given its effectiveness and efficiency, we hope that \Ours\ can serve as a strong and simple alternative of current mainstream anchor-based detectors. We also believe that \Ours\ can be extended to solve many other instance-level recognition tasks.

\section*{Appendix}

\section{Class-agnostic Precision-recall Curves}
\begin{table}[!h]
\centering 
\small
\begin{tabular}{ l| c | c c c }
Method & AP & AP$_{50}$ & AP$_{75}$ & AP$_{90}$ \\
\Xhline{2\arrayrulewidth}
Orginal RetinaNet \cite{lin2017focal} & 39.5 & 63.6 & 41.8 & 10.6 \\
RetinaNet w/ GN \cite{wu2018group} & 40.0 & 64.5 & 42.2 & 10.4 \\
\Ours & \textbf{40.5} & \textbf{64.7}  & \textbf{42.6} & \textbf{13.1} \\
\hline
& & +0.2 & +0.4 & +2.7
\end{tabular}
\vspace{0.2cm}
\caption{The class-agnostic detection performance for RetinaNet and \Ours. \Ours\ has better performance than RetinaNet. Moreover, the improvement over RetinaNet becomes larger with a stricter IOU threshold. The results are obtained with the same models in Table 4 of our main paper.}
\label{table:detection_performance}
\end{table}

\begin{figure}[t!]
\begin{center}
\includegraphics[width=\linewidth]{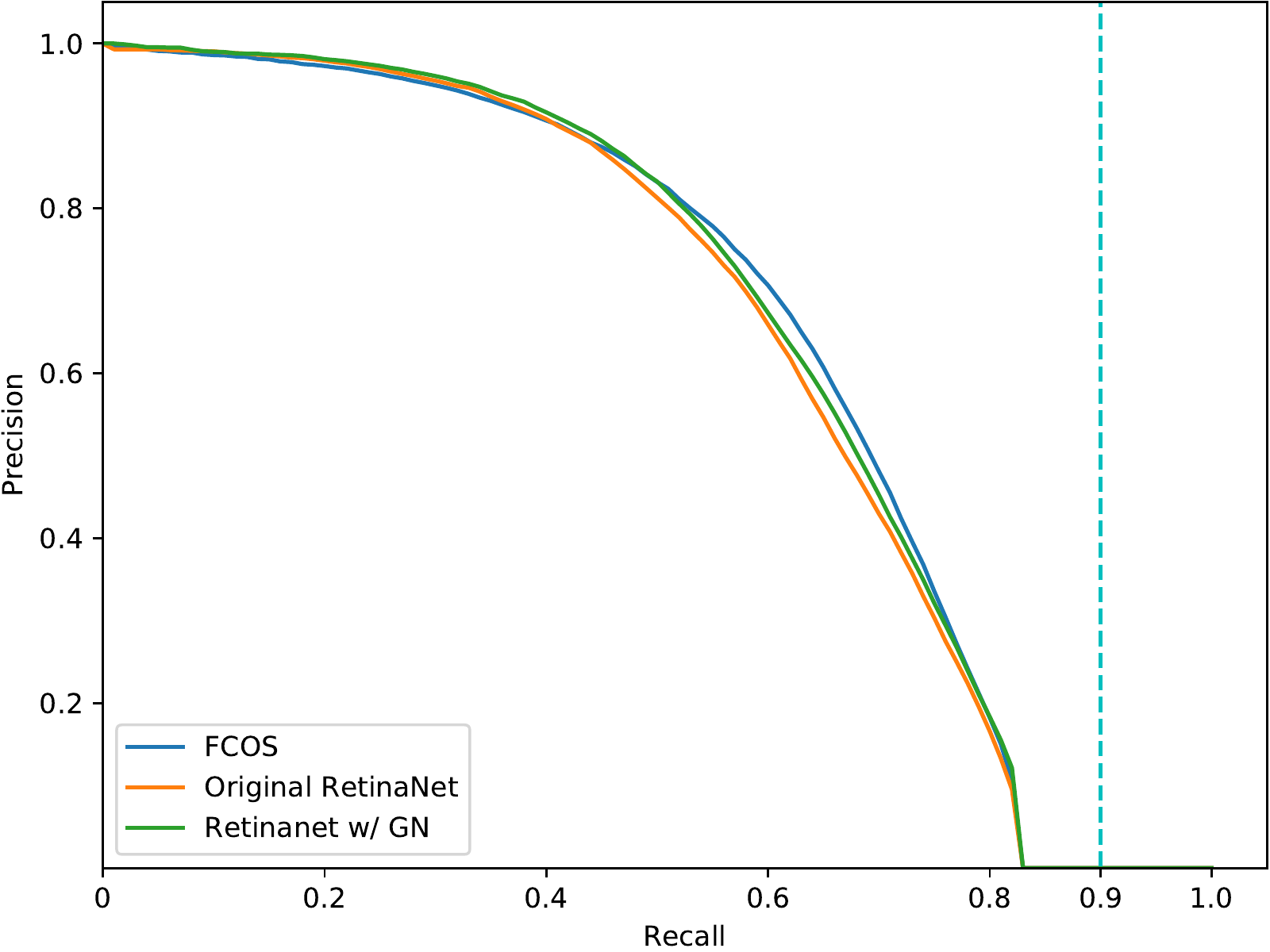}
\end{center}
   \caption{Class-agnostic precision-recall curves at IOU $= 0.50$.}
\label{fig:pr_iou_50}
\end{figure}

\begin{figure}[t!]
\begin{center}
\includegraphics[width=\linewidth]{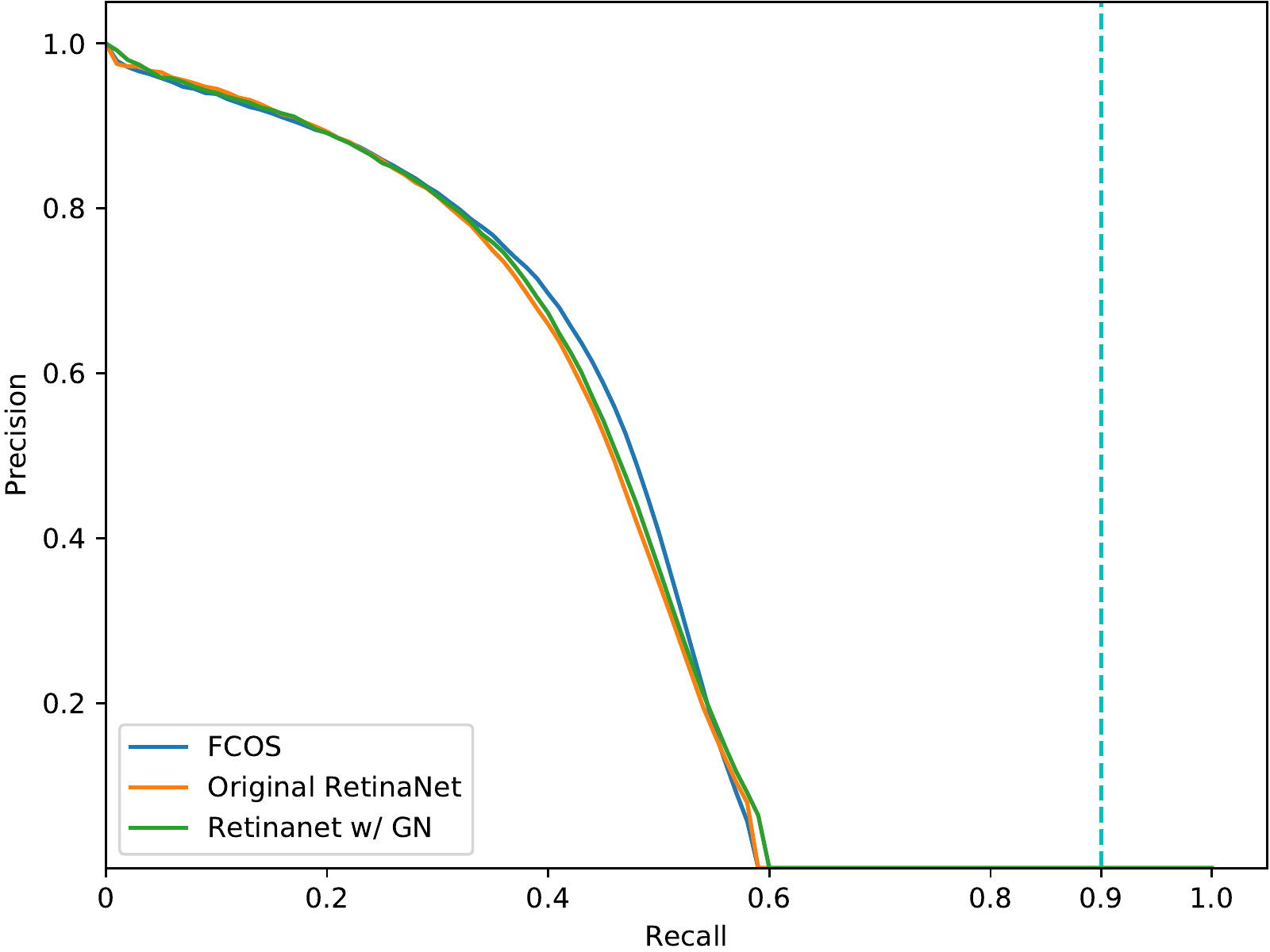}
\end{center}
   \caption{Class-agnostic precision-recall curves at IOU $= 0.75$.}
\label{fig:pr_iou_75}
\end{figure}

\begin{figure}
\begin{center}
\includegraphics[width=\linewidth]{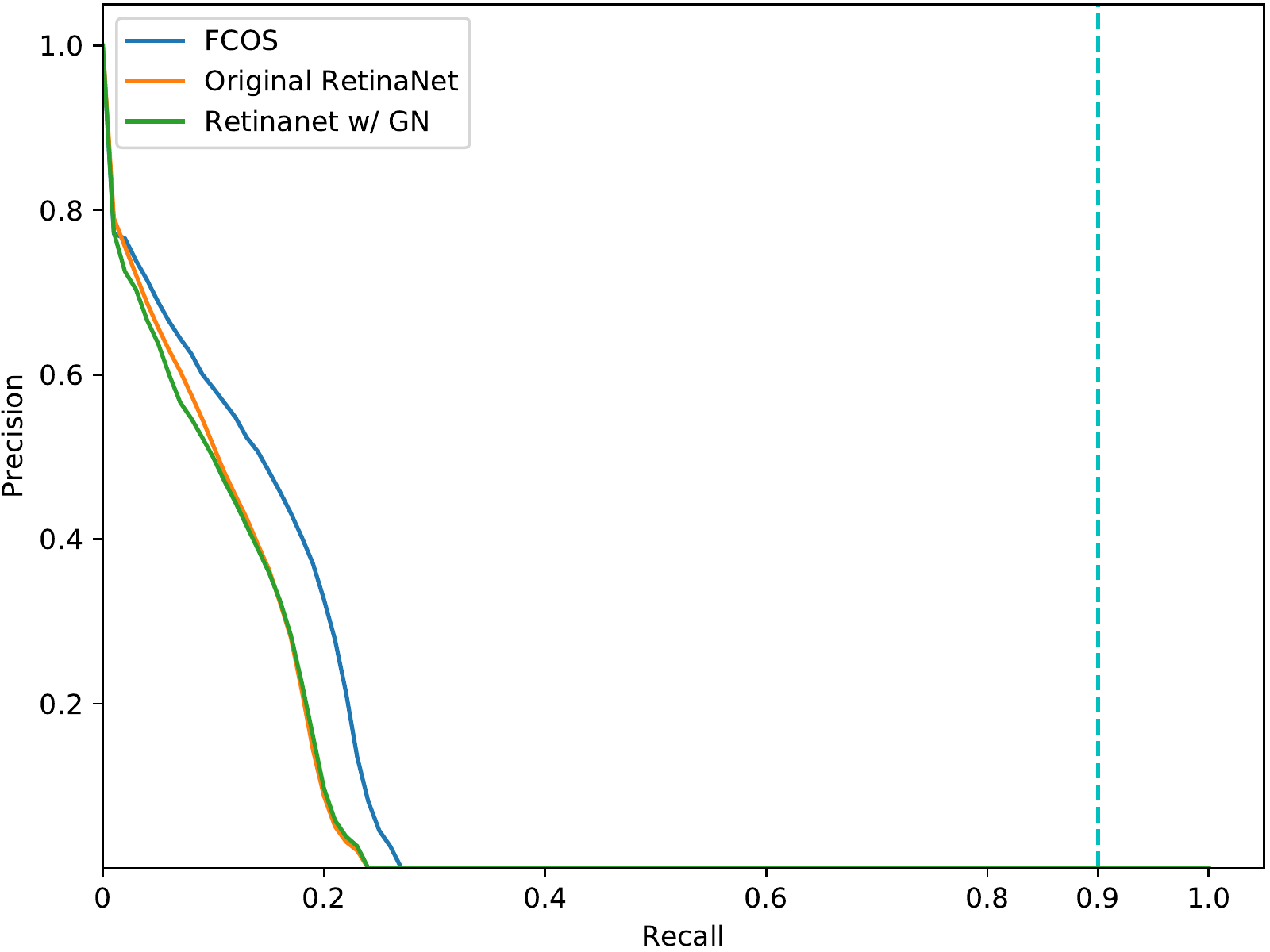}
\end{center}
   \caption{Class-agnostic precision-recall curves at IOU $= 0.90$.}
\label{fig:pr_iou_90}
\end{figure}

In Fig.~\ref{fig:pr_iou_50}, Fig.~\ref{fig:pr_iou_75} and Fig.~\ref{fig:pr_iou_90}, we present class-agnostic precision-recall curves on split $\tt minival$
at IOU thresholds being 0.50, 0.75 and 0.90, respectively. Table \ref{table:detection_performance} shows APs corresponding to the three curves.

As shown in Table \ref{table:detection_performance}, our \Ours\ achieves better performance than its anchor-based counterpart RetinaNet. Moreover, it worth noting that with a stricter IOU threshold, \Ours\ enjoys a larger improvement over RetinaNet, which suggests that \Ours\ has a better bounding box regressor to detect objects more accurately. One of the reasons should be that \Ours\ has the ability to leverage more foreground samples to train the regressor as mentioned in our main paper.

Finally, as shown in all precision-recall curves, the best recalls of these detectors in the precision-recall curves are much lower than $90\%$. It further suggests that the small gap ($98.40\%$ vs. $99.23\%$) of \textit{best possible recall (BPR)} between \Ours\ and RetinaNet hardly harms the final detection performance.

\section{Visualization for Center-ness}
\begin{figure*}[t!]
\begin{center}
    \includegraphics[width=.8\linewidth]{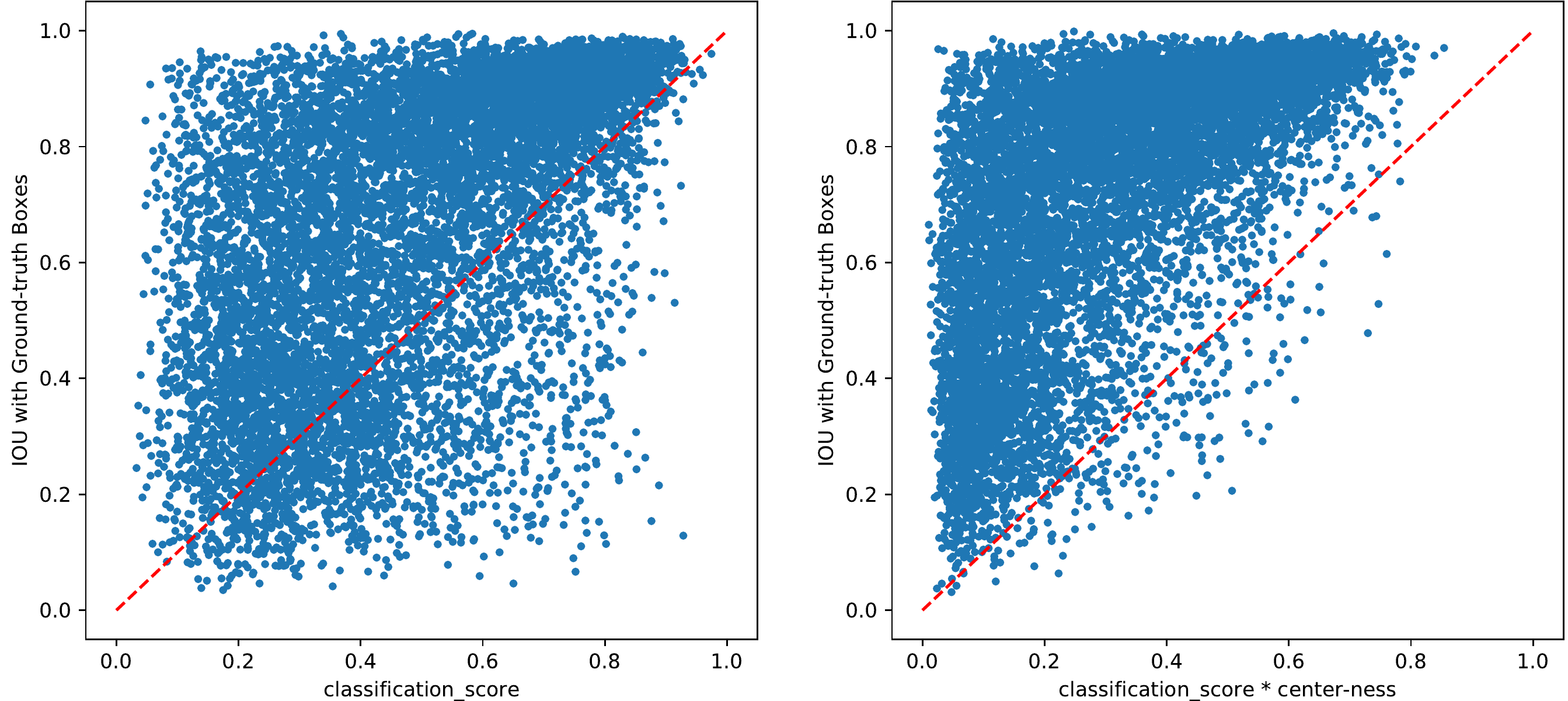}
\end{center}
   \caption{Without (left) or with (right) the proposed center-ness. A point in the figure denotes a detected bounding box. The dashed line is the line $y = x$. As shown in the figure (right), after multiplying the classification scores with the center-ness scores, the low-quality boxes (under the line $y=x$) are pushed to the left side of the plot. It suggests that the scores of these boxes are reduced substantially.}
\label{fig:centerness}
\end{figure*}
As mentioned in our main paper, by suppressing low-quality detected bounding boxes, the proposed center-ness branch improves the detection performance by a large margin. In this section, we confirm this. 

We expect that the center-ness can down-weight the scores of low-quality bounding boxes such that these bounding boxes can be filtered out in following post-processing such as non-maximum suppression (NMS). A detected bounding box is considered as a low-quality one if it has a low IOU score with its corresponding ground-truth bounding box. 
A bounding box with low IOU but  a high confidence score  is likely to become a false positive and harm the precision.

In Fig.~\ref{fig:centerness}, we consider a detected bounding box as a 2D point $(x, y)$ with $x$ being its score and $y$ being the IOU with its corresponding ground-truth box. As shown in Fig.~\ref{fig:centerness} (left), before applying the center-ness, there are a large number of low-quality bounding boxes but with a high confidence score (i.e., the points under the line $y=x$). Due to their high scores, these low-quality bounding boxes cannot be eliminated in post-processing and result in 
lowering the precision of the detector.
After multiplying the classification score with the center-ness score, these points are pushed to %
the left side of the plot
(i.e., their scores are reduced), as shown in Fig.~\ref{fig:centerness} (right). As a result, these low-quality bounding boxes are much more likely to be filtered out in post-processing and the final detection performance can be improved.

\section{Qualitative Results}
\begin{figure*}
\begin{center}
\includegraphics[width=\linewidth]{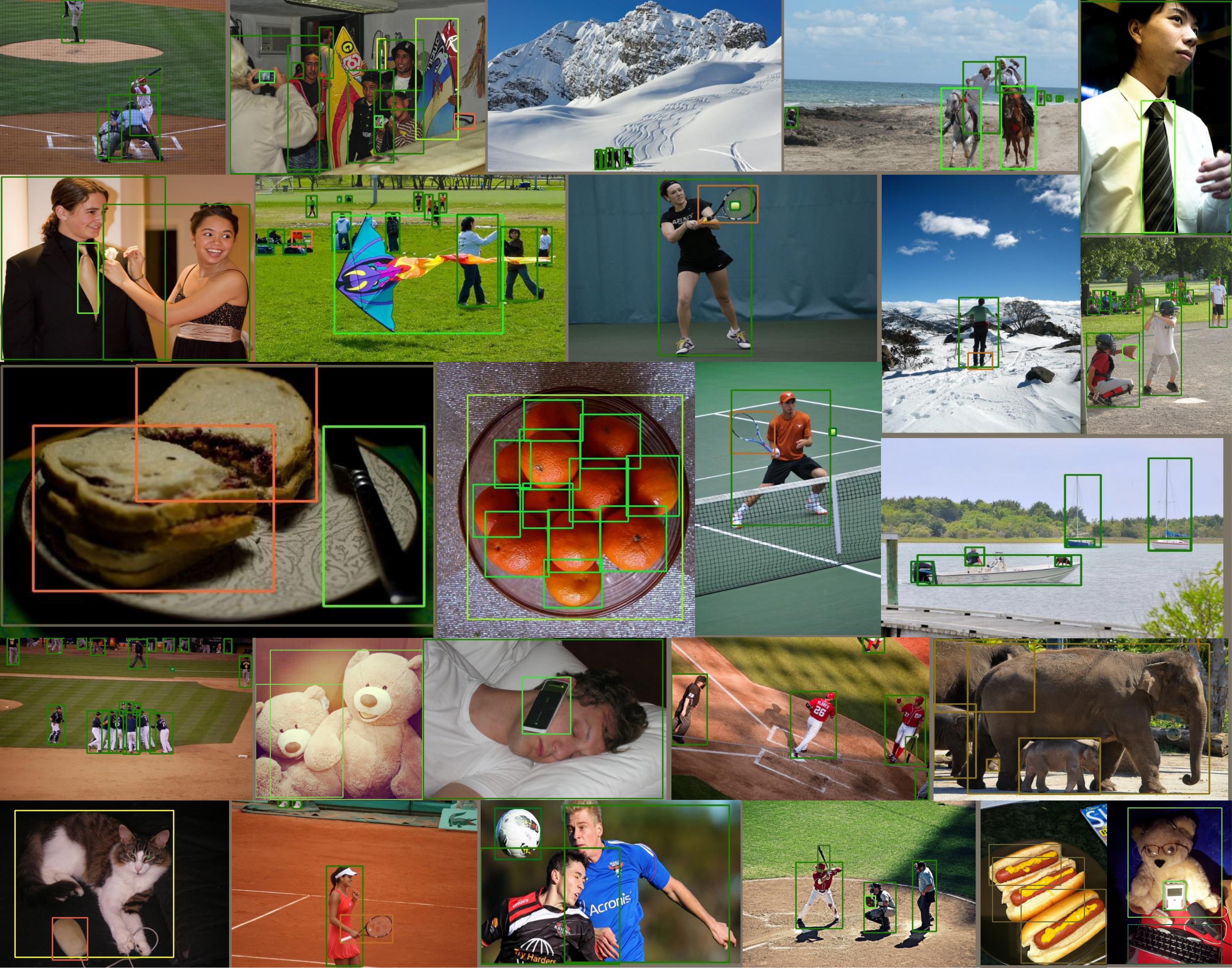}
\end{center}
   \caption{Some detection results on $\tt minival$ split. ResNet-50 is used as the backbone. As shown in the figure, \Ours\ works well with a wide range of objects including crowded, occluded, highly overlapped, extremely small and very large objects.}
\label{fig:visualization_results}
\end{figure*}
Some qualitative results are shown in Fig. \ref{fig:visualization_results}. As shown in the figure, our proposed \Ours\ can detect a wide range of objects including crowded, occluded, highly overlapped, extremely small and very large objects.

\def\Ours{{FCOS}\xspace}
\def\eg{{\it e.g.}\xspace}
\def\ie{{\it i.e.}\xspace}

\section{More discussions}
\noindent\textbf{Center-ness vs.\ IoUNet:}

 Center-ness and IoUNet of Jiang et al.\ ``Acquisition of Localization Confidence for Accurate Object Detection''
  shares a similar purpose (\ie, to suppress low-quality predictions) with different approaches.
  IoUNet trains a separate network to predict the IoU score between predicted bounding-boxes and ground-truth boxes. Center-ness, as a part of our
   detector, only has a single layer and is trained jointly with the detector, thus being much simpler. Moreover, ``center-ness'' does not take as input the predicted bounding-boxes. Instead, it directly accesses the location's ability to predict high-quality bounding-boxes.

\noindent\textbf{BPR in Section 4.1 and ambiguity analysis:}

 We do not aim to compare ``recall by specific IoU" with ``recall by pixel within box".
 The main purpose of Table 1 is to show that the upper bound of recall of \Ours is very close to the upper bound of recall of anchor-based RetinaNet (98.4\%  vs.\ 99.23\%).
 BPR by other IoU thresholds are listed as those
  are used in the official code of RetinaNet. Moreover, \textit{no evidence shows that the regression targets of \Ours are difficult to learn because they are more spread-out.}  \Ours in fact yields more accurate bounding-boxes.

 During training, we deal with the ambiguity at the same FPN level by choosing the ground-truth
  box with the minimal area. When testing, if two objects A and B with the same class have overlap,
  no matter which one object the locations in the overlap predict, the prediction is correct and the missed one can be predicted by the locations only belonging to it. In the case that A and B do not belong to the same class,
  a location in the overlap might predict A's class but regress B's bounding-box, which is a mistake.
  That is why we only count the ambiguity across different classes. Moreover,
it appears that
  {this ambiguity does not make \Ours worse than RetinaNet in AP, as shown in Table \ref{table:ablation_study}.}


\noindent\textbf{Additional ablation study:}

\noindent
 As shown in Table \ref{table:ablation_study}, a  vanilla
  \Ours performs on par with RetinaNet, being of
   simpler design and with $\sim9\times$ less network outputs. Moreover, \Ours works much better than RetinaNet with single anchor.
   As for the $2\%$ gain on $\tt test$-$\tt dev$, besides the performance gain brought by the components in Table \ref{table:ablation_study}, we conjecture that different training details (\eg, learning rate schedule) might
cause slight differences in performance.

\begin{table}[t!]
\centering
\small
\begin{tabular}{l | c c c c | c}
Method & $C_5/P_5$ & GN & Scalar & IoU & AP \\
\Xhline{2\arrayrulewidth}
RetinaNet (\#A=1) & $C_5$ & & & & 32.5 \\
RetinaNet (\#A=9) & $C_5$ & & & & 35.7 \\
\hline
\Ours (pure) & $C_5$ & & & & 35.7 \\
\Ours & $P_5$ & & & & 35.8 \\
\Ours & $P_5$ & \checkmark & & & 36.3 \\
\Ours & $P_5$ & \checkmark & \checkmark & & 36.4 \\
\Ours & $P_5$ & \checkmark & \checkmark & \checkmark & \textbf{36.6} \\
\end{tabular}
\caption{Ablation study on MS-COCO \texttt{minival}. ``\#A" is the number of anchor boxes per location in RetinaNet. ``IOU" is IOU loss. ``Scalar" denotes whether to use scalars in $\exp$. All experiments are conducted with the same settings.}
\label{table:ablation_study}
\end{table}

\noindent\textbf{RetinaNet with Center-ness: 
}

 Center-ness cannot be directly used in RetinaNet with multiple anchor boxes per location because one location on feature maps has \textit{only one} center-ness score but different anchor boxes on the same location require different ``center-ness"
 (note that center-ness is also used as ``soft" thresholds for positive/negative samples).

  For anchor-based RetinaNet, the IoU score between anchor boxes and ground-truth boxes may serve as an alternative of ``center-ness".



\noindent\textbf{Positive samples overlap with RetinaNet:}

 We want to highlight that center-ness comes into play \textit{only} when testing. When training, all locations within ground-truth
  boxes are marked as positive samples. As a result, \Ours can use more foreground locations to train the regressor and thus yield more accurate bounding-boxes.

\paragraph{Acknowledgments.} We would like to thank the author of \cite{fcosplus} for %
the tricks of 
center sampling and GIoU. We also thank Chaorui Deng for HRNet based \Ours\ and his suggestion of positioning the center-ness branch with box regression.

{\small
 \bibliographystyle{ieee_fullname}
\bibliography{FCOS}

\begin{thebibliography}{10}\itemsep=-1pt

\bibitem{fcosplus}
\url{https://github.com/yqyao/FCOS_PLUS}, 2019.

\bibitem{boominathan2016crowdnet}
Lokesh Boominathan, Srinivas~SS Kruthiventi, and R~Venkatesh Babu.
\newblock Crowdnet: A deep convolutional network for dense crowd counting.
\newblock In {\em Proc. ACM Int. Conf. Multimedia}, pages 640--644. ACM, 2016.

\bibitem{ICCV2017Chen}
Yu Chen, Chunhua Shen, Xiu-Shen Wei, Lingqiao Liu, and Jian Yang.
\newblock Adversarial {PoseNet}: A structure-aware convolutional network for
  human pose estimation.
\newblock In {\em Proc. IEEE Int. Conf. Comp. Vis.}, 2017.

\bibitem{deng2009imagenet}
Jia Deng, Wei Dong, Richard Socher, Li-Jia Li, Kai Li, and Li Fei-Fei.
\newblock {ImageNet}: A large-scale hierarchical image database.
\newblock In {\em Proc. IEEE Conf. Comp. Vis. Patt. Recogn.}, pages 248--255.
  IEEE, 2009.

\bibitem{fu2017dssd}
Cheng-Yang Fu, Wei Liu, Ananth Ranga, Ambrish Tyagi, and Alexander Berg.
\newblock {DSSD}: Deconvolutional single shot detector.
\newblock {\em arXiv preprint arXiv:1701.06659}, 2017.

\bibitem{girshick2015fast}
Ross Girshick.
\newblock Fast {R-CNN}.
\newblock In {\em Proc. IEEE Conf. Comp. Vis. Patt. Recogn.}, pages 1440--1448,
  2015.

\bibitem{Detectron2018}
Ross Girshick, Ilija Radosavovic, Georgia Gkioxari, Piotr Doll\'{a}r, and
  Kaiming He.
\newblock Detectron.
\newblock \url{https://github.com/facebookresearch/detectron}, 2018.

\bibitem{he2016deep}
Kaiming He, Xiangyu Zhang, Shaoqing Ren, and Jian Sun.
\newblock Deep residual learning for image recognition.
\newblock In {\em Proc. IEEE Conf. Comp. Vis. Patt. Recogn.}, pages 770--778,
  2016.

\bibitem{He_2019_CVPR}
Tong He, Chunhua Shen, Zhi Tian, Dong Gong, Changming Sun, and Youliang Yan.
\newblock Knowledge adaptation for efficient semantic segmentation.
\newblock In {\em Proc. IEEE Conf. Comp. Vis. Patt. Recogn.}, June 2019.

\bibitem{he2018end}
Tong He, Zhi Tian, Weilin Huang, Chunhua Shen, Yu Qiao, and Changming Sun.
\newblock An end-to-end textspotter with explicit alignment and attention.
\newblock In {\em Proc. IEEE Conf. Comp. Vis. Patt. Recogn.}, pages 5020--5029,
  2018.

\bibitem{huang2017speed}
Jonathan Huang, Vivek Rathod, Chen Sun, Menglong Zhu, Anoop Korattikara,
  Alireza Fathi, Ian Fischer, Zbigniew Wojna, Yang Song, Sergio Guadarrama,
  et~al.
\newblock Speed/accuracy trade-offs for modern convolutional object detectors.
\newblock In {\em Proc. IEEE Conf. Comp. Vis. Patt. Recogn.}, pages 7310--7311,
  2017.

\bibitem{huang2015densebox}
Lichao Huang, Yi Yang, Yafeng Deng, and Yinan Yu.
\newblock Densebox: Unifying landmark localization with end to end object
  detection.
\newblock {\em arXiv preprint arXiv:1509.04874}, 2015.

\bibitem{law2018cornernet}
Hei Law and Jia Deng.
\newblock Cornernet: Detecting objects as paired keypoints.
\newblock In {\em Proc. Eur. Conf. Comp. Vis.}, pages 734--750, 2018.

\bibitem{lin2017feature}
Tsung-Yi Lin, Piotr Doll{\'a}r, Ross Girshick, Kaiming He, Bharath Hariharan,
  and Serge Belongie.
\newblock Feature pyramid networks for object detection.
\newblock In {\em Proc. IEEE Conf. Comp. Vis. Patt. Recogn.}, pages 2117--2125,
  2017.

\bibitem{lin2017focal}
Tsung-Yi Lin, Priya Goyal, Ross Girshick, Kaiming He, and Piotr Doll{\'a}r.
\newblock Focal loss for dense object detection.
\newblock In {\em Proc. IEEE Conf. Comp. Vis. Patt. Recogn.}, pages 2980--2988,
  2017.

\bibitem{lin2014microsoft}
Tsung-Yi Lin, Michael Maire, Serge Belongie, James Hays, Pietro Perona, Deva
  Ramanan, Piotr Doll{\'a}r, and Lawrence Zitnick.
\newblock Microsoft {COCO}: Common objects in context.
\newblock In {\em Proc. Eur. Conf. Comp. Vis.}, pages 740--755. Springer, 2014.

\bibitem{Depth2015Liu}
Fayao Liu, Chunhua Shen, Guosheng Lin, and Ian Reid.
\newblock Learning depth from single monocular images using deep convolutional
  neural fields.
\newblock {\em {IEEE} Trans. Pattern Anal. Mach. Intell.}, 2016.

\bibitem{liu2016ssd}
Wei Liu, Dragomir Anguelov, Dumitru Erhan, Christian Szegedy, Scott Reed,
  Cheng-Yang Fu, and Alexander~C Berg.
\newblock {SSD}: Single shot multibox detector.
\newblock In {\em Proc. Eur. Conf. Comp. Vis.}, pages 21--37. Springer, 2016.

\bibitem{Liu_2019_CVPR}
Yifan Liu, Ke Chen, Chris Liu, Zengchang Qin, Zhenbo Luo, and Jingdong Wang.
\newblock Structured knowledge distillation for semantic segmentation.
\newblock In {\em Proc. IEEE Conf. Comp. Vis. Patt. Recogn.}, June 2019.

\bibitem{long2015fully}
Jonathan Long, Evan Shelhamer, and Trevor Darrell.
\newblock Fully convolutional networks for semantic segmentation.
\newblock In {\em Proc. IEEE Conf. Comp. Vis. Patt. Recogn.}, pages 3431--3440,
  2015.

\bibitem{redmon2016you}
Joseph Redmon, Santosh Divvala, Ross Girshick, and Ali Farhadi.
\newblock You only look once: Unified, real-time object detection.
\newblock In {\em Proc. IEEE Conf. Comp. Vis. Patt. Recogn.}, pages 779--788,
  2016.

\bibitem{redmon2017yolo9000}
Joseph Redmon and Ali Farhadi.
\newblock {YOLO9000}: better, faster, stronger.
\newblock In {\em Proc. IEEE Conf. Comp. Vis. Patt. Recogn.}, pages 7263--7271,
  2017.

\bibitem{redmon2018yolov3}
Joseph Redmon and Ali Farhadi.
\newblock Yolov3: An incremental improvement.
\newblock {\em arXiv preprint arXiv:1804.02767}, 2018.

\bibitem{ren2015faster}
Shaoqing Ren, Kaiming He, Ross Girshick, and Jian Sun.
\newblock {Faster R-CNN}: Towards real-time object detection with region
  proposal networks.
\newblock In {\em Proc. Adv. Neural Inf. Process. Syst.}, pages 91--99, 2015.

\bibitem{shrivastava2016beyond}
Abhinav Shrivastava, Rahul Sukthankar, Jitendra Malik, and Abhinav Gupta.
\newblock Beyond skip connections: Top-down modulation for object detection.
\newblock In {\em Proc. IEEE Conf. Comp. Vis. Patt. Recogn.}, 2017.

\bibitem{DBLP:journals/corr/abs-1902-09212}
Ke Sun, Bin Xiao, Dong Liu, and Jingdong Wang.
\newblock Deep high-resolution representation learning for human pose
  estimation.
\newblock In {\em Proc. IEEE Conf. Comp. Vis. Patt. Recogn.}, 2019.

\bibitem{szegedy2017inception}
Christian Szegedy, Sergey Ioffe, Vincent Vanhoucke, and Alexander~A Alemi.
\newblock Inception-v4, inception-resnet and the impact of residual connections
  on learning.
\newblock In {\em Proc. National Conf. Artificial Intell.}, 2017.

\bibitem{tian2019decoders}
Zhi Tian, Tong He, Chunhua Shen, and Youliang Yan.
\newblock Decoders matter for semantic segmentation: Data-dependent decoding
  enables flexible feature aggregation.
\newblock In {\em Proc. IEEE Conf. Comp. Vis. Patt. Recogn.}, pages 3126--3135,
  2019.

\bibitem{wu2018group}
Yuxin Wu and Kaiming He.
\newblock Group normalization.
\newblock In {\em Proc. Eur. Conf. Comp. Vis.}, pages 3--19, 2018.

\bibitem{xie2017aggregated}
Saining Xie, Ross Girshick, Piotr Doll{\'a}r, Zhuowen Tu, and Kaiming He.
\newblock Aggregated residual transformations for deep neural networks.
\newblock In {\em Proc. IEEE Conf. Comp. Vis. Patt. Recogn.}, pages 1492--1500,
  2017.

\bibitem{Yin2019enforcing}
Wei Yin, Yifan Liu, Chunhua Shen, and Youliang Yan.
\newblock Enforcing geometric constraints of virtual normal for depth
  prediction.
\newblock In {\em Proc. IEEE Int. Conf. Comp. Vis.}, 2019.

\bibitem{yu2016unitbox}
Jiahui Yu, Yuning Jiang, Zhangyang Wang, Zhimin Cao, and Thomas Huang.
\newblock Unitbox: An advanced object detection network.
\newblock In {\em Proc. ACM Int. Conf. Multimedia}, pages 516--520. ACM, 2016.

\bibitem{zhou2017east}
Xinyu Zhou, Cong Yao, He Wen, Yuzhi Wang, Shuchang Zhou, Weiran He, and Jiajun
  Liang.
\newblock {EAST}: an efficient and accurate scene text detector.
\newblock In {\em Proc. IEEE Conf. Comp. Vis. Patt. Recogn.}, pages 5551--5560,
  2017.

\bibitem{zhu2019feature}
Chenchen Zhu, Yihui He, and Marios Savvides.
\newblock Feature selective anchor-free module for single-shot object
  detection.
\newblock In {\em Proc. IEEE Conf. Comp. Vis. Patt. Recogn.}, June 2019.

\end{thebibliography}
}

\end{document}